\pdfoutput=1

\documentclass[11pt]{article}

\usepackage{acl}

\usepackage{times}
\usepackage{latexsym}
\usepackage{graphicx}
\usepackage{todonotes}
\usepackage{amsmath,amssymb}
\usepackage{url}
\usepackage{color,soul}
\usepackage{makecell} 
\usepackage{multirow}
\usepackage{tikz}
\usepackage{enumitem}
\usepackage{caption, subcaption}

\usepackage[T1]{fontenc}

\usepackage[utf8]{inputenc}

\usepackage{microtype}

\title{Multilingual and Multimodal Topic Modelling with Pretrained Embeddings}

\author{
  Elaine Zosa \and Lidia Pivovarova \\
  Department of Computer Science\\
  University of Helsinki\\
  Helsinki, Finland\\
  \texttt{firstname.lastname@helsinki.fi} \\ 
}

\begin{document}
\maketitle
\begin{abstract}

This paper presents M3L-Contrast---a novel multimodal multilingual (M3L) neural topic model for comparable data that maps texts from multiple languages and images into a shared topic space. Our model is trained jointly on texts and images and takes advantage of pretrained document and image embeddings to abstract the complexities between different languages and modalities. As a multilingual topic model, it produces aligned language-specific topics and as multimodal model, it infers textual representations of semantic concepts in images. We demonstrate that our model is competitive with a zero-shot topic model in predicting topic distributions for comparable multilingual data and significantly outperforms a zero-shot model in predicting topic distributions for comparable texts and images. We also show that our model performs almost as well on unaligned embeddings as it does on aligned embeddings.

\end{abstract}
\section{Introduction}
Topic modelling is an unsupervised method initially designed for text data that extracts latent themes in documents through the co-occurrence statistics of the words in the documents. In most probabilistic topic models, a topic is a distribution over a vocabulary and a document is a distribution over topics~\cite{blei2003latent}. Multilingual topic models extend basic topic models for multilingual data by jointly training on multiple languages~\cite{mimno2009polylingual,hao2018learning}. These models learn aligned language-specific topics and have been used in different cross-lingual applications such as multilingual news clustering~\cite{de2009cross} and comparing discourses from different cultures in news and social media~\cite{shi2016detecting,gutierrez2016detecting}.

Most topic models are designed for textual data but there is also a rich body of work on applying topic modelling to images resulting in multimodal topic models~\cite{barnard2003matching,feng2010visual,roller2013multimodal}. These models use natural language supervision to improve the semantic representation of images. Augmenting topical information from text with topic information from visual inputs also produces better semantic representations of words.

Neural topic models have been proposed to improve on classical topic models and have resulted in models that are more computationally efficient and produces more coherent topics~\cite{srivastava2017autoencoding}. Moreover, the neural topic modelling framework has given rise to models that take advantage of information from external sources such as word embeddings~\cite{dieng2020topic} and contextualised language models~\cite{bianchi-etal-2021-pre,bianchi-etal-2021-cross,hoyle2020improving,mueller2021fine}.

In this work, we present a novel neural multilingual \textit{and} multimodal topic model that takes advantage of pretrained document and image embeddings to abstract the complexities between languages and modalities. Our work is based on the contextualized topic model~\cite[CTM,][]{bianchi-etal-2021-pre,bianchi-etal-2021-cross}, a family of topic models that uses contextualized document embeddings as input. 

We show that while ZeroshotTM~\cite{bianchi-etal-2021-cross}, a cross-lingual variant of CTM, can predict relevant topic distributions of documents in languages it has not seen during training, this ability does not transfer well to unseen modalities (e.g. images). Moreover, since ZeroshotTM only sees monolingual data, it produces monolingual topics that are inferred from documents in a single language. This approach does not take into account possible biases in worldviews that are hidden in different languages. 

Our approach, which we refer to as M3L-Contrast, trains jointly on multilingual texts and images with a contrastive objective. We show that our model produces better topic distributions for comparable texts and images compared to ZeroshotTM even with unaligned embeddings and our model also improves on a classical multilingual topic model for comparable multilingual data. The main contributions of this work are:
\begin{enumerate}[nosep]
    \item  We present a neural multimodal \textit{and} multilingual topic model for comparable data that maps images and texts into a shared topic space;
    \item we show that contrastive learning is effective in mapping embeddings from unaligned encoders into a shared topic space \textit{and} improves on the alignment of aligned embeddings; 
    \item we present a multilingual topic model for comparable multilingual data that uses pretrained embeddings and improves on a classical topic model for comparable data.\footnote{Our code is available at \url{https://github.com/ezosa/M3L-topic-model}}
\end{enumerate}

\section{Related Work}
\subsection{Neural topic models}
Neural topic models (NTMs) refer to a class of topic models that use neural networks to estimate the parameters of the topic-word and document-topic distributions. Using a variational autoencoder (VAE) to map documents into latent topic spaces was proposed by \citet{srivastava2017autoencoding} and demonstrated in the ProdLDA model that exhibited better topic coherences and faster training than classical models. This has led to other VAE-based topic models that can incorporate information from external sources such as the Embedded Topic Model~\cite{dieng2020topic} which uses pretrained word embeddings, the Contextualised Topic Model~\cite{bianchi-etal-2021-pre} which uses contextualised embeddings and the BERT-based Autoencoder as Teacher~\cite{hoyle2020improving} model that distills large language models to improve topic coherence. 

\subsection{Multilingual topic models}
Multilingual topic models infer aligned language-specific topics from a multilingual dataset. To align topics across languages, some degree of supervision is required to establish the link between the languages. In most cases, the languages are linked either at a word level or at a document level~\cite{hao2020empirical}. Models that use word-level supervision require a translation dictionary to link words from different languages~\cite{jagarlamudi2010extracting,hao2018learning,yang2019multilingual}. Document-level supervision requires a comparable dataset where a document in one language is linked to a thematically similar document in another language~\cite{mimno2009polylingual,de2009cross}. 

The Polylingual Topic Model \cite[PLTM,][]{mimno2009polylingual} is widely-used classical multilingual topic model for comparable data. To our knowledge, the Neural Multilingual Topic Model~\cite{wu2020learning}, a model that uses word-level supervision, is the only neural multilingual topic model so far. ZeroshotTM~\cite{bianchi-etal-2021-cross}, while not a multilingual model, is capable of zero-shot cross-lingual topic inference: it can predict topic distributions for documents in unseen languages if the model is trained on embeddings from a multilingual encoder.  However, ZeroshotTM requires \textit{aligned} embeddings for zero-shot topic modelling.

\subsection{Multimodal topic models} 
Multimodal topic models use data from different modalities 
to infer topics. The most popular pairing is texts and images. 
Some text-and-image topic models use labelled image datasets to learn natural language representations of images using a supervised topic modelling approach~\cite{barnard2003matching,zheng2014topic}. Other models extract `visual words' from images using image feature extractors such as SIFT and images are represented as a bag of `visual words' in the same manner that documents are represented as a bag of textual words~\cite{feng2010visual,virtanen2012factorized,roller2013multimodal}.
\cite{an2020multimodal} trained visual and textual topic models from neural network representations for multimodal depression detection but does not map text and images into the same topic space.

\subsection{Contrastive learning} 
Contrastive learning is a self-supervised technique that uses different views of the same data to learn better data representations~\cite{jaiswal2021survey,liu2021self}. In contrastive training the goal is to minimize the distance between positive samples while separating them from negative samples.  Contrastive training is popular in multimodal settings such as web-scale text-image alignment~\cite{radford2021learning,jia2021scaling}, audio-visual alignment~\cite{khorrami2021evaluation} and biomedical imaging~\cite{zhang2020contrastive}.

In neural topic modelling, contrastive learning has recently been used to improve on the Adversarial Topic Model~\cite{wang2019atm} by adding a contrastive objective to the training loss and taking a more principled approach to sampling positive and negative samples~\cite{nguyen2021contrastive}. 

\section{Multilingual and Multimodal Model}
\subsection{Neural multilingual topic model}
\label{sec:models-multilingual}

\begin{figure}[t!]
    \centering
    \includegraphics[width=0.8\columnwidth]{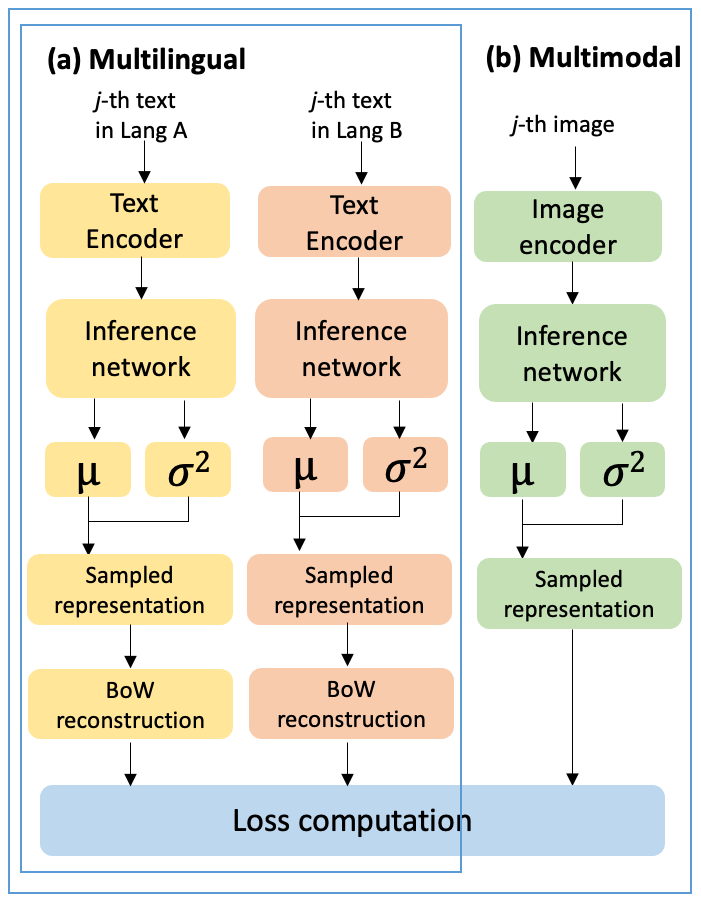}
    \caption{Proposed M3L-Contrast topic model. (a) Multilingual topic model with language-specific encoders and inference networks; (b) Extension to the multimodal setting. The loss function is detailed in Equation~\ref{eq:loss-m3l-contrast}.}
    \label{fig:archi-m3l-contrast}
\end{figure}

We first propose a neural multilingual topic model for comparable multilingual data that uses pretrained document embeddings. Our multilingual model is based on ZeroshotTM~\cite{bianchi-etal-2021-cross}, a zero-shot cross-lingual topic model. However, we are not aiming for a zero-shot model. Instead, our model infers aligned \textit{language-specific topics} for each language present in the dataset. Moreover, our approach does not require the pretrained document embeddings to be aligned beforehand. This property makes it advantageous in settings where a multilingual encoder that includes our desired language might not exist such as in low-resource settings.

Figure~\ref{fig:archi-m3l-contrast}(a) shows the multilingual model architecture. The model uses independent inference networks for each language. To align language-specific topics, the model minimizes the Kullback-Leibler (KL) divergence between the topic distributions of comparable documents from different languages and, in addition, uses a contrastive loss to map similar instances close to each other in the topic space and keep non-related instances apart.  

For each tuple of aligned documents in the comparable multilingual dataset, we encode the documents from each language using their own separate encoders (whether aligned or non-aligned) and then the embeddings are passed to language-specific inference networks that infers the mean, $\mu$, and variance, $\sigma^2$, of the Gaussian distribution from which we sample latent document-topic distributions. At this point, the languages are independent of each other and have not yet shared any information.  

After sampling topic distributions for each document, we induce a shared topic space by minimizing the pairwise KL divergence between the language-specific distributions whose parameters are estimated from their own inference network. We also add a contrastive objective so that aligned examples are kept away from other examples in the topic space. 
We use InfoNCE~\cite{van2018representation} as our contrastive loss. The positive pairs are all possible combinations of document pairs from the same tuple and negative pairs are all other pairs of documents from different tuples within a batch. For instance, for a comparable dataset with two languages and batch size $N$, we would have $N$ positive pairs and $N^2-N$ negative pairs per batch. For three languages, that would be $3N$ positive pairs and $3(N^2-N)$ negative pairs, etc.

Thus, the loss consists of the three components: the reconstruction loss; the KL divergence between topic distributions; and the contrastive loss. 
Formally, the loss function is written as:
\begin{align}
\mathcal{L} & = \sum_{l=0}^{L} \mathbb{E}_{q}[w^{\top} \log(\text{softmax}(\beta_{l}\theta_{d}))] - \nonumber \\
& \sum_{\substack{a,b=0\\a \neq b}}^{n} \mathbb{KL}(p(\theta_{i}^{a} | x_{i}^{a}) || q(\theta_{i}^{b} | x_{i}^{b})) - \nonumber \\ 
& s \sum_{\substack{a,b=0\\a \neq b}}^{n} \log \frac{\exp( (\theta_{i}^{a} \cdot \theta_{i}^{b}) / \tau)}{ \sum_{j=0}^{N} \sum_{c,d=0}^{n} \exp( (\theta_{i}^{c} \cdot \theta_{j}^{d}) / \tau)}    
\label{eq:loss-m3l-contrast}
\end{align} 

The first term is the sum of the bag-of-words (BoW) reconstruction losses of each language in the corpus. 
We refer the reader to~\cite{srivastava2017autoencoding} for further details on the reconstruction loss.

The second term is the sum of the KL divergences between the language-specific document distributions, $p()$ and $q()$, whose mean and variance are estimated from language-specific inference networks; $\theta$ refers to the sampled topic representation of a document in a tuple where $i$ is the tuple index, $a$ and $b$ are the indices of the documents inside the tuple and $n$ is the size of the tuple. Lastly, $x$ refers to a document embedding.    

The third term is the InfoNCE loss where $(\theta_{i}^{a} \cdot \theta_{i}^{b})$ are positive pairs (they belong to the same tuple) and $(\theta_{i}^{c} \cdot \theta_{j}^{d})$ are negative pairs (they are from different tuples). $N$ is the batch size, $\tau$ is the temperature and $s$ is a constant to give additional weight to the contrastive loss.

\subsection{Extension to multimodal setting}
\label{sec:models-multimodal}
We now extend the proposed multilingual topic model to the \textit{multimodal} setting. Figure~\ref{fig:archi-m3l-contrast}(b) shows the architecture of the proposed multilingual \textit{and} multimodal topic model.  

We can think of the multimodal case as a generalization of the multilingual model. The loss function in Equation~\ref{eq:loss-m3l-contrast} remains essentially the same. Since a BoW representation is not available for images, the reconstruction loss is computed only on texts and the first loss term is unchanged. In the second term of the loss function, $x$ can be a document \textit{or} image embedding and $\theta$ is the sampled topic distribution for that embedding. 

Since the document or image embeddings abstract the modality of the data, the topic distributions are now modality-agnostic. Thus, the third term is also unchanged, except for the tuple size $n$. A multimodal dataset with one language and one image view would have $N$ positive and $N^2-N$ negative pairs, the same as in the bilingual case. For two languages \textit{and} one image, we would have $3N$ positive pairs and $3(N^2-N)$ negative pairs, as in the trilingual case. 

We refer to our proposed topic model as \textbf{M3L-Contrast} for \textit{multimodal multilingual (M3L) topic model with contrastive learning}.

\section{Experimental Setup}
\subsection{Dataset}
We run experiments on our proposed model on a dataset of aligned English and German Wikipedia articles and images. We take aligned articles from the Wikipedia Comparable Corpora\footnote{\url{https://linguatools.org/tools/corpora/wikipedia-comparable-corpora/}} and align them with images from the Wikipedia-based Image Text dataset (WIT)~\cite{srinivasan2021wit} \footnote{\url{https://github.com/google-research-datasets/wit}}. We use articles instead of the image descriptions in WIT because topic models are designed for full documents, rather than snippets of text. 

We randomly select 20,000 tuples for training. Since articles can be associated with more than one image and we want fixed-size tuples during training, we randomly select one image per article pair. For testing, we randomly select 1,000 article pairs. We consider \textit{all} images aligned with the paired articles, which results in 3,278 unique tuples in the test set.  

\subsection{Evaluation}
We evaluate M3L-Contrast in the multilingual setting and the multimodal setting, separately. In the multilingual case we train M3L-Contrast on multilingual articles \textit{without} images and for the multimodal case on the multilingual articles \textit{and} images.

We want document-topic distributions for multilingual articles and images from the same tuple to be similar to each other 
\textit{and} distinct from other examples. Thus, we evaluate the alignment of topic distributions using retrieval tasks.\footnote{We are aware that topic distributions do not outperform raw embeddings in retrieval tasks but the point of this evaluation is not to improve cross-lingual or cross-modal retrieval but to evaluate the alignment of the topic distributions.}

Texts and images are fed one at a time to their own language-specific and modality-specific inference networks to obtain topic distributions. For the multilingual setting, we match an English article to the most similar German article in terms of the Jensen-Shannon divergence (JSD) between their respective document-topic distributions. For the multimodal setting, we match English articles to images and German articles to images, separately. We use mean reciprocal rank (MRR) to measure text retrieval performance and uninterpolated average precision~\cite[UAP,][]{manning1999topics} to measure text-image retrieval performance because multiple images can be associated with an article.

We also report the averaged JSD between the topic distributions for all data pairs from the same tuple. Lastly, we compute language-specific topic coherences with respect to the training data using normalised pointwise mutual information~\cite[NPMI,][]{roder2015exploring}\footnote{Computed using the Gensim library~\cite{rehurek2010software}.}. 

\subsection{Baselines}
    ~~~\textbf{PLTM}~\cite{mimno2009polylingual} We implement PLTM with Gibbs sampling.
 
    \textbf{ZeroshotTM}~\cite{bianchi-etal-2021-cross} We train separate models on the English and German articles using the authors' original implementation.\footnote{\url{https://github.com/MilaNLProc/contextualized-topic-models}}
 
    \textbf{ZeroshotTM-KD}~\cite{pivovarova2021visual} We adapt ZeroshotTM for multilingual or multimodal settings using knowledge distillation~\cite{hinton2015distilling}. This method uses the parameters learned by the teacher model as priors for the student. We train four separate teacher-student pairs: (1) Model trained on English articles as teacher, German as student; (2) German articles as teacher, English as student; (3) English articles as teacher, images as student; and (4) German articles as teacher, images as student.

\subsection{Configurations}
We report the performance of the neural topic models using CLIP~\cite{radford2021learning} as our multimodal multilingual encoder for a fair comparison.\footnote{ \textit{clip-ViT-B-32} for images and  \textit{clip-ViT-B-32-multilingual-v1} for texts.} CLIP is a pretrained vision-language model trained on web-scale data that encodes text and images into a common embedding space. We train all models with 100 topics for 100 epochs. Other hyperparameters are discussed in the Appendix. We use batch size 32 for M3L-Contrast. In Section~\ref{sec:ablation-study} we show the performance of M3L-Contrast for different encoder combinations (aligned and unaligned), different batch sizes and topic numbers.



\section{Results and Discussion}

\subsection{Multilingual setting}

\begin{table}[t!]
\centering
\resizebox{\columnwidth}{!}{  
\begin{tabular}{l | r r | r r}
\hline
{} & \multicolumn{2}{c}{\textbf{Coherence$\uparrow$}} & \multicolumn{2}{c}{\textbf{EN-DE text}}\\
\textbf{Model} & \textbf{EN} &  \textbf{DE} & \textbf{MRR$\uparrow$} & \textbf{JSD$\downarrow$}\\
\hline 
PLTM & {0.064} & {0.044} & {0.333} & {0.067}\\
\textbf{ZeroshotTM}  & {0.113} & {0.096} & \textbf{0.997} & \textbf{0.012}\\
ZeroshotTM-KD & 0.109 & {0.092} & {0.390} & {0.081}\\ 
\hline \hline
M3L-Contrast & \textbf{0.119} & \textbf{0.097} & {0.684} & {0.036}\\ 
\end{tabular}
}
\caption{Language-specific topic coherences (NPMI) and cross-lingual retrieval peformance (MRR and JSD).}
\label{tab:results-text-only}
\end{table}

Table~\ref{tab:results-text-only} shows the cross-lingual retrieval performance and averaged JSD of aligned articles. ZeroshotTM is the clear winner with an MRR of 0.997 and the lowest JSD. M3L-Contrast, while it does not outperform ZeroshotTM, shows encouraging results given that it has to infer twice as many topics as ZeroshotTM (bilingual case). It also outperforms PLTM, a classical multilingual topic model and the only other model, aside from M3L-Contrast, trained on multilingual articles. Moreover, M3L-Contrast also has the best topic coherences.

\subsection{Multimodal setting}

\begin{table*}[h!]
\centering
\begin{tabular}{l | r r | r r | r r }
\hline
{} & \multicolumn{2}{c}{\textbf{Coherence$\uparrow$}}  & \multicolumn{2}{c}{\textbf{EN-images}} & \multicolumn{2}{c}{\textbf{DE-images}} \\
\textbf{Model} & \textbf{EN} & \textbf{DE} & \textbf{UAP$\uparrow$} & \textbf{JSD$\downarrow$}  & \textbf{UAP$\uparrow$} & \textbf{JSD$\downarrow$}\\
\hline 
ZeroshotTM & {0.113} & {0.096} & {0.034} & {0.445} & {0.039} & {0.435}\\
ZeroshotTM-KD & {0.109} & {0.092} & {0.082} & {0.128} & {0.093} & {0.146}\\
\hline \hline
\textbf{M3L-Contrast} & \textbf{0.122} & \textbf{0.097} & \textbf{0.125} & \textbf{0.130} & \textbf{0.102} & \textbf{0.147}\\
\end{tabular}
\caption{Language-specific topic coherences (NPMI) and text-image retrieval performance (UAP). JSD is the averaged JS divergence between topic distributions of aligned articles and images. Only M3L-Contrast is jointly trained on \textbf{multilingual articles and images}.}
\label{tab:results-text-images}
\end{table*}

Table~\ref{tab:results-text-images} shows the results for text-image matching. M3L-Contrast performs the best with UAP of 0.125 and 0.102 for matching English and German articles to images, respectively, and has the lowest JSDs. In a reversal of the results for cross-lingual retrieval, ZeroshotTM performs the worst with the lowest UAP scores and highest JSDs. ZeroshotTM-KD only slightly outperforms ZeroshotTM, indicating that the success of M3L-Contrast can be attributed to \textit{joint} training and cannot be achieved with the teacher-student sequential training scheme.

These results indicate that ZeroshotTM without any modifications is not suitable for multimodal settings. One likely reason is that multimodal encoders like CLIP suffer from the so-called `modality gap' where embeddings for different modalities are mapped to separate regions in the embedding space \cite{ModalityGap2022}. 

Our results also indicate that for a joint multimodal and multilingual neural topic model, it could be beneficial to use a hybrid model that uses  separate inference networks for different modalities and a shared network for the same modality. We leave this for future work.

\subsection{Error analysis}
\begin{table*}[ht!]
\begin{center}
\begin{tabular}{r p{2.8cm} | p{2.2cm} | p{2.3cm} | p{2.3cm}}
    \hline
    {} & {} & \textbf{EN article} & \textbf{DE article} & \textbf{Image}\\
    \hline\hline
    \multicolumn{5}{c}{Article title: \textbf{Capsicum pubescens}} \\
    \hline
    \multirow{6}{*}{\includegraphics[width=0.2\textwidth]{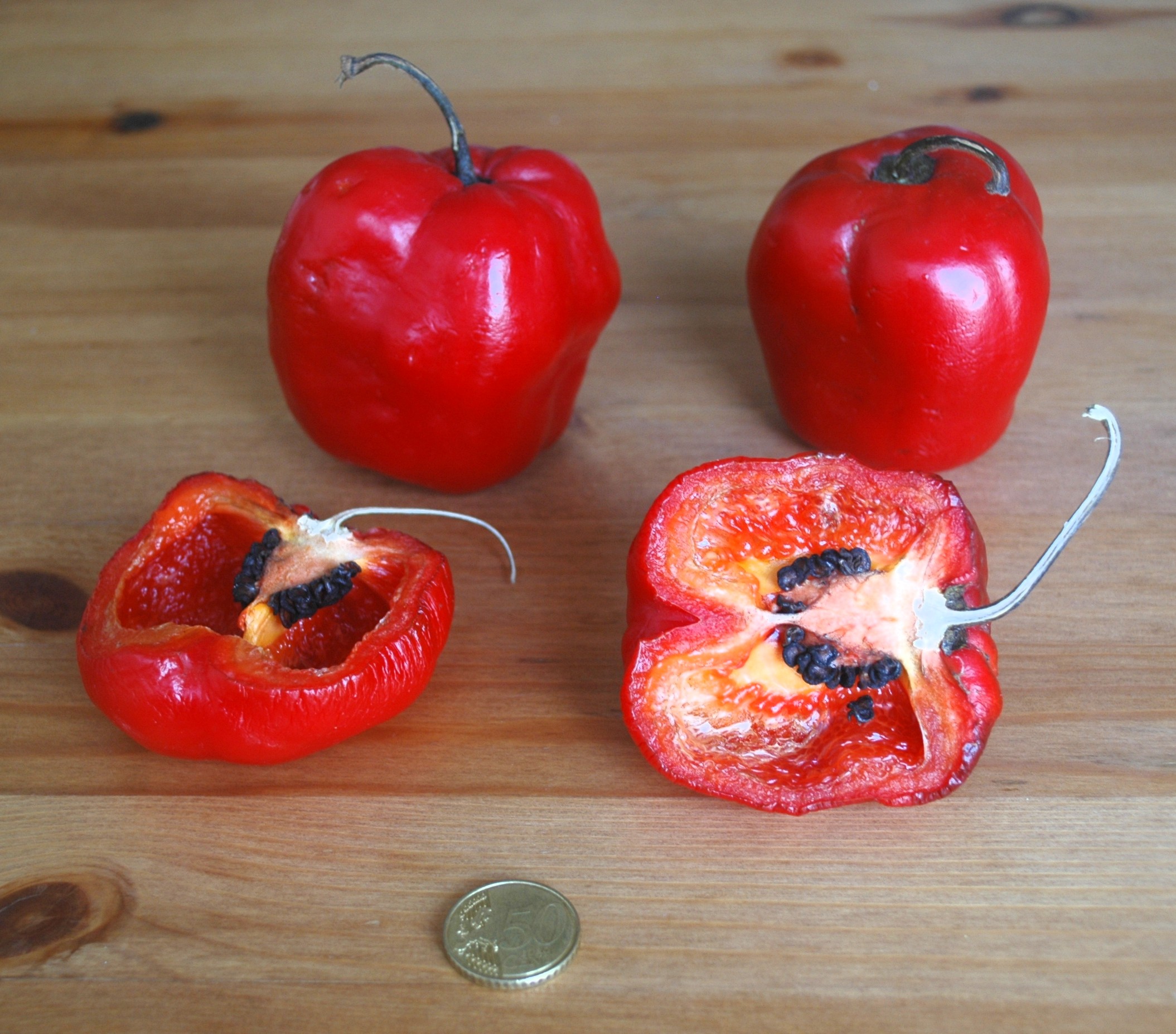}} & \textbf{ZeroshotTM} & {\textbf{21:} plant, leaves, flowers, tall} & {\textbf{31:} bird, south, america, species} & {\textbf{67:} university, library, museum, research}\\
    & \textbf{\textbf{ZeroshotTM-KD} \textit{(EN teacher)}} & \textbf{44: } plant, leaves, flowers, plants, genus & - & \textbf{44: } plant, leaves, flowers, plants, genus \\
    & \textbf{M3L-Contrast} & {\textbf{65:} plant, plants, leaves, flowers} 
    & {\textbf{65:} beschreibung (\textit{description}), pflanzen (\textit{plant}), selten 
    (\textit{rare}), stehen (\textit{stand})}
    &{\textbf{65:} plant, plants, leaves, flowers}\\
    \hline
    \hline
    \multicolumn{5}{c}{Article title: \textbf{Microexpression}/\textbf{Mikroexpression}} \\
    \hline
    \multirow{5}{*}{\includegraphics[width=0.25\textwidth]{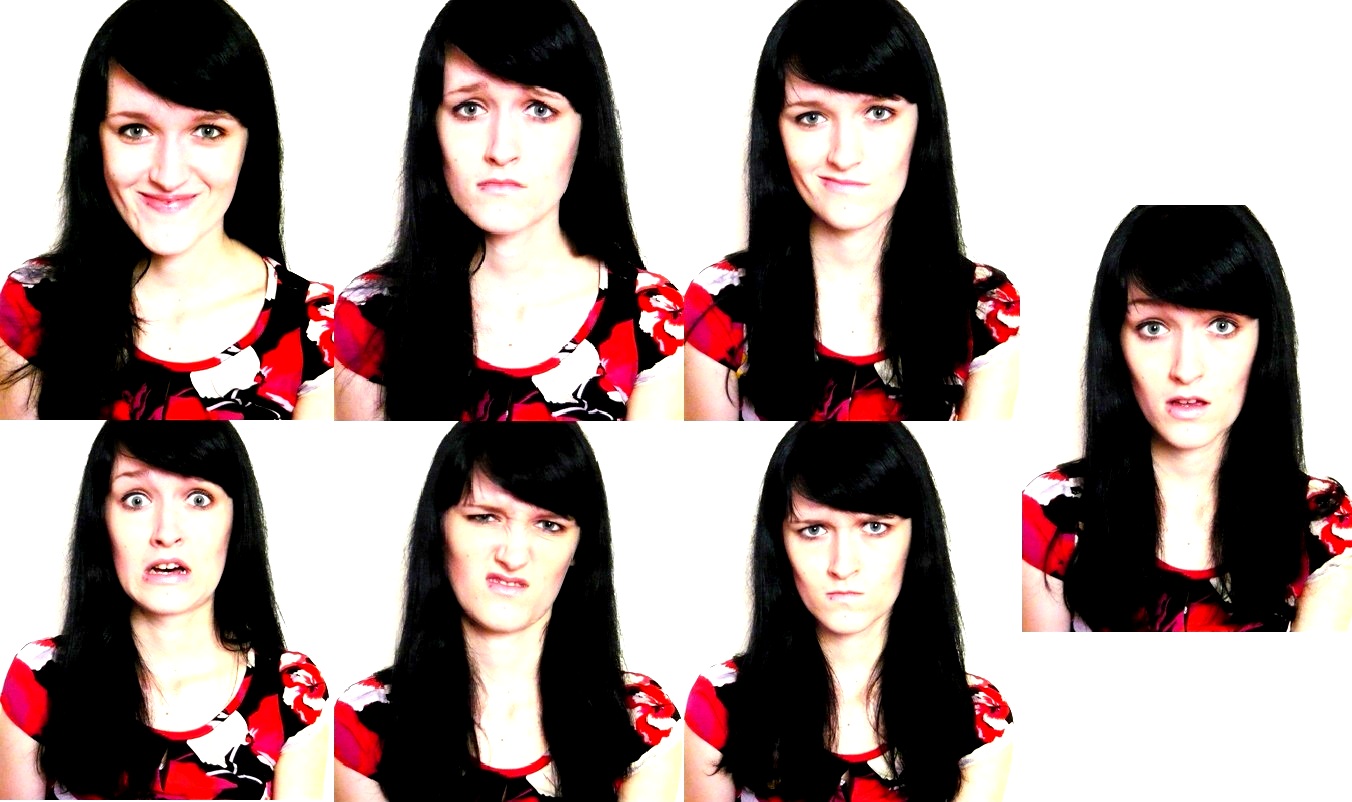}} & \textbf{ZeroshotTM} & {\textbf{13}: include, cause, may, cases, occur} & {\textbf{13:} include, cause, may, cases, occur} & {\textbf{9:} bishop, catholic, pope, church, roman}\\
     & \textbf{\textbf{ZeroshotTM-KD} \textit{(EN teacher)}} & \textbf{32:} blood, symptoms, disease, cell, bone & - & \textbf{69:} album, released, song, single, group \\
    & \textbf{M3L-Contrast} & {\textbf{84:} theory, term, example, social, defined} & {\textbf{84:} begriff (\textit{concept}), definition (\textit{definition}), beispiel (\textit{example}), theorie (\textit{theory}), zahl (\textit{number})} & {\textbf{5:} film, award, series, actress, born}\\
    \hline
\end{tabular}
\caption{Top topics of Wikipedia article pairs and a related image. The numbers indicate the topic indices.}
\label{tab:examples}
\end{center}
\end{table*}
To further investigate differences between the models we checked some examples from the test set. We show two article-image tuples and their predicted topics in~Table~\ref{tab:examples}. The table contains the top topic for the aligned English and German articles (titles shown) and an image associated with them. 

The first example article is about pepper\footnote{\url{https://en.wikipedia.org/wiki/Capsicum_pubescens}}. ZeroshotTM predicts relevant topics for the English and German articles but off-topic for the image. 
ZeroshotTM-KD (a teacher model trained on English articles and a student on images) predicts a relevant topic for the English article and the image but it has not been trained on  German. M3L-Contrast predicts relevant topics for the English and German articles and the image. 
Though the table shows English topic labels for the image, it is equally possible to produce German image labels.

In the second example, the article about microexpressions\footnote{\url{https://en.wikipedia.org/wiki/Microexpression}} is illustrated with an image of a woman presenting basic emotions. ZeroshotTM predicts slightly relevant topics for the English and German articles but off-topic for the image. ZeroshotTM-KD also predicts a relevant topic for the English article. Although the image topic is different from the topic of the article, it is still somewhat relevant in that the model may have associated images of women with pop stars. M3L-Contrast predicts relevant topics for the English and German articles. For the image, it predicts a topic about actresses likely because the image depicts a woman.

We found similar behaviour in other cases: English and German articles are usually assigned with the same topic while the image often has a different topic. In many cases M3L-Contrast finds an aligned topic for an image while the other two models fail.


\subsection{Visualizing the topic space}

\begin{figure*}[t!]
     \centering
     \begin{subfigure}{\textwidth}
        \centering
        \includegraphics[width=0.7\textwidth]{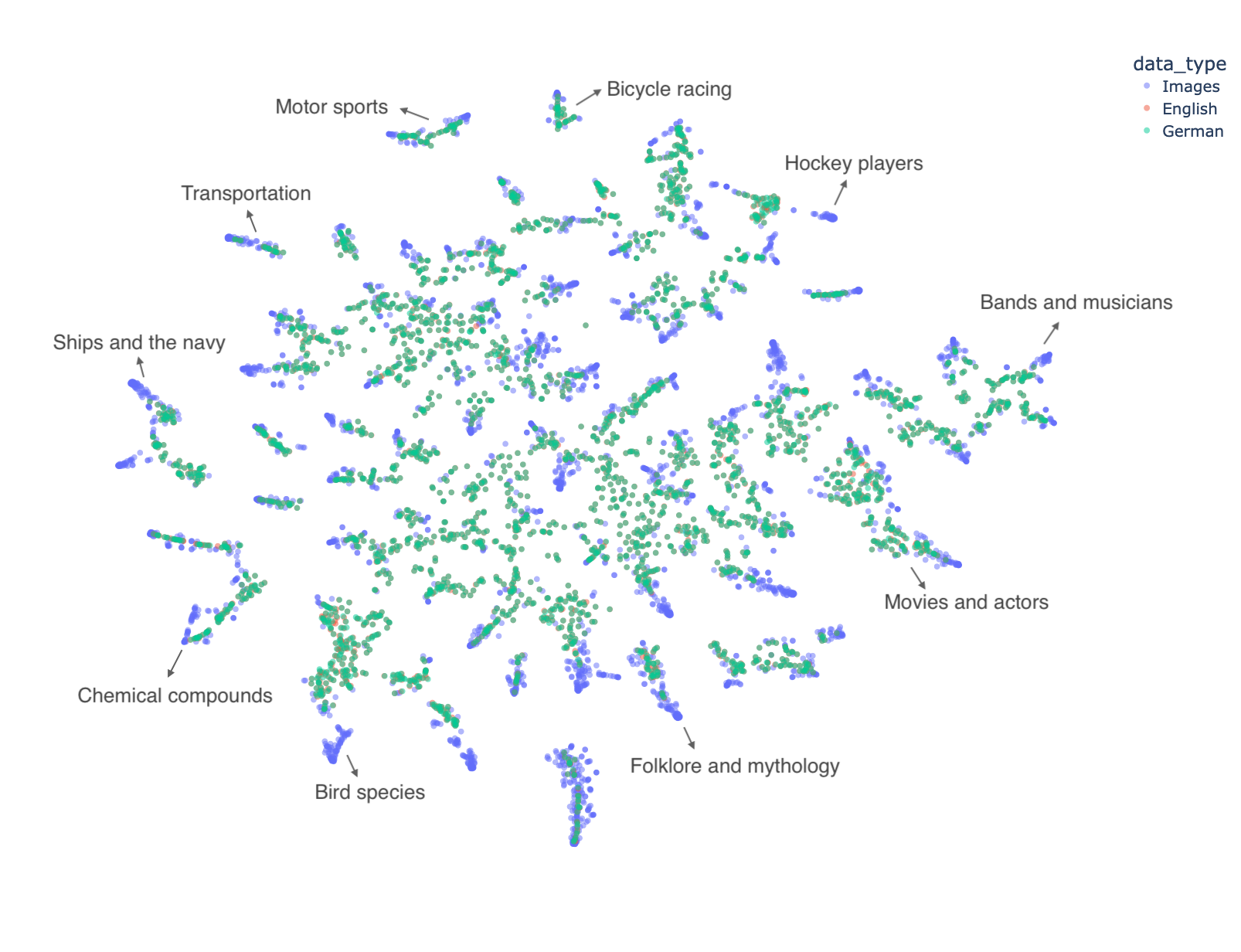}
        \caption{ZeroshotTM}
        \label{fig:topic-space-ctm}
     \end{subfigure}
     \begin{subfigure}{\textwidth}
        \centering
        \includegraphics[width=0.7\textwidth]{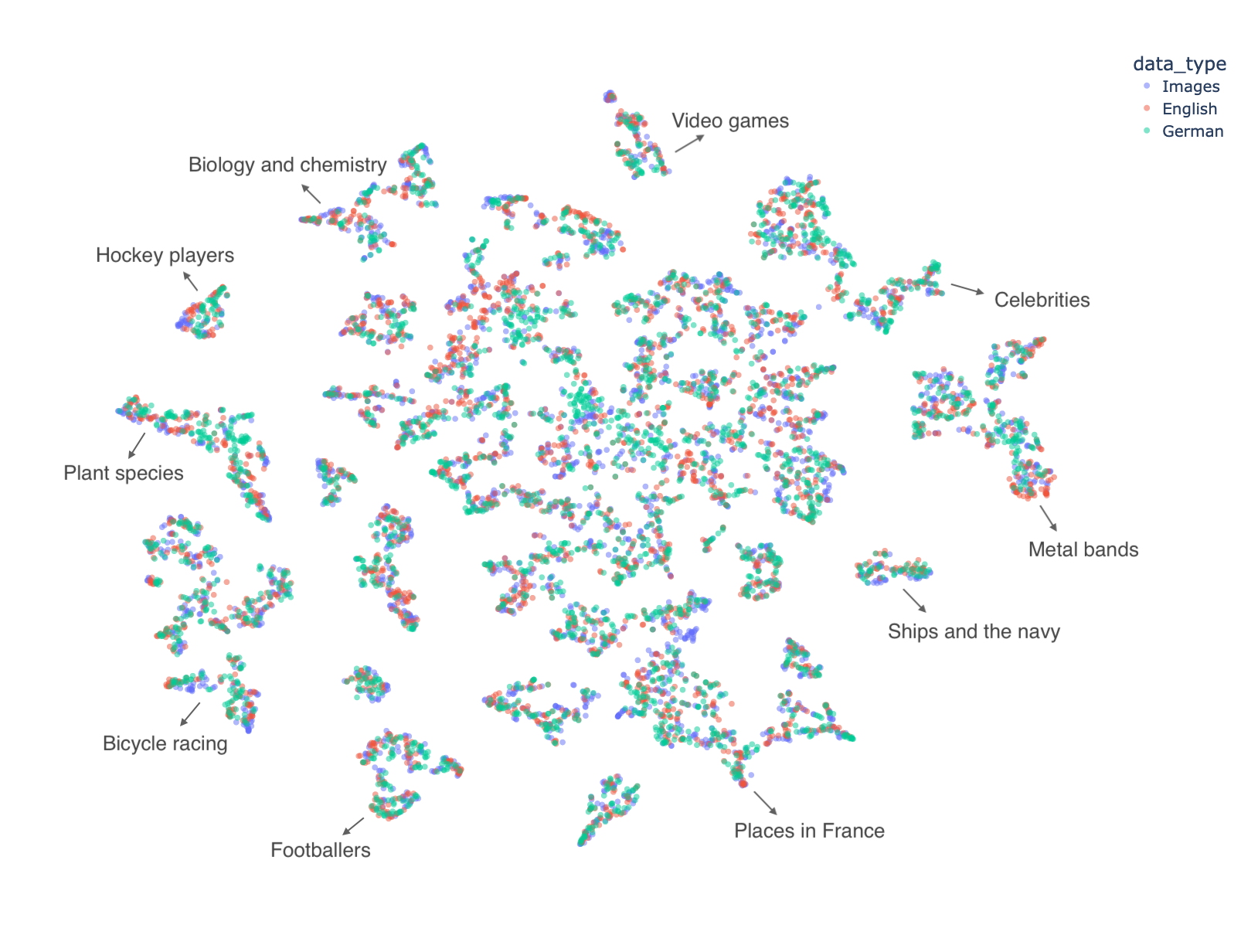}
        \caption{M3L-Contrast}
        \label{fig:topic-space-m3l-contrast}
     \end{subfigure}
    \caption{tSNE visualizations of topic distributions of multilingual texts and images inferred by ZeroshotTM and M3L-Contrast, respectively. Annotations are added manually. Best viewed in color.}
    \label{fig:topic-spaces}
\end{figure*}

To investigate the structure of the multimodal multilingual topic space, we use 2D visualizations presented in Figures~\ref{fig:topic-space-ctm} and~\ref{fig:topic-space-m3l-contrast}, for ZeroshotTM and M3L-Contrast, respectively. These figures show the proximities of multilingual texts and images, represented by their predicted topic distributions, in the topic spaces induced by the respective models and mapped into two dimensions with tSNE.\footnote{These figures are available as interactive plots in the code repository of this paper.}

In Figure~\ref{fig:topic-space-ctm}---the ZeroshotTM topic space---the topic distributions of the aligned articles are very similar to their counterparts (most of the points representing English articles are hidden under the German articles). The images, however, tend to be isolated instead of being close to their textual counterparts. This supports the modality gap hypothesis and explains why ZeroshotTM performs poorly in the text-image retrieval task.

In Figure~\ref{fig:topic-space-m3l-contrast}, the topic space induced by M3L-Contrast, articles and images tend to group together in terms of \textit{themes}---exactly the behaviour we want from a topic model. No single modality or language is isolated by itself. This explains why M3L-Contrast performs better than ZeroshotTM in text-image retrieval. On the other hand, the English and German articles are not as close to each other as in ZeroshotTM. This supports our claim that joint training takes into account data from all languages and adjusts for possible discrepancies between worldviews across languages, even though this property results in worse performance in cross-lingual text retrieval.

\section{Ablation Study}
\label{sec:ablation-study}

\begin{figure}[t!]
    \centering
    \includegraphics[width=0.5\textwidth]{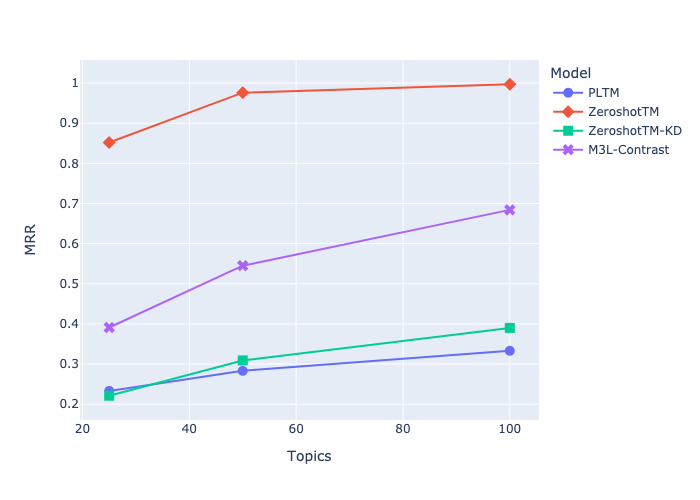}
    \caption{Effect of increasing topic numbers on cross-lingual retrieval performance (MRR).}
    \label{fig:text-only-topic-numbers}
\end{figure}

\begin{figure}[t!]
    \centering
    \includegraphics[width=0.5\textwidth]{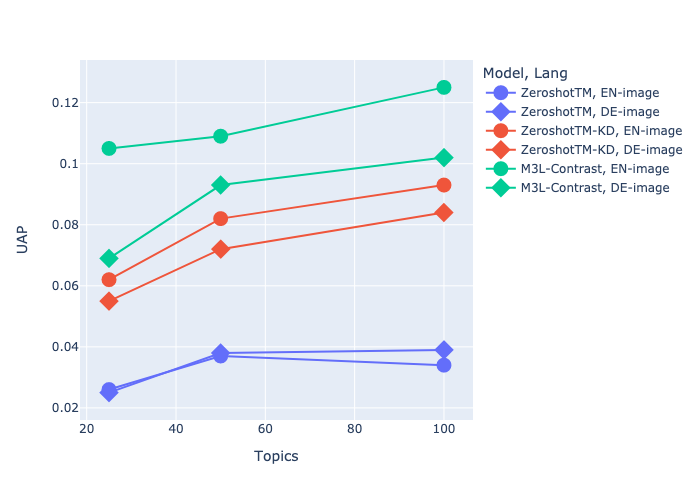}
    \caption{Effect of increasing topic numbers on text-image retrieval performance (UAP).}
    \label{fig:text-images-topic-numbers}
\end{figure}

\begin{figure}[t!]
    \centering
    \includegraphics[width=0.55\textwidth]{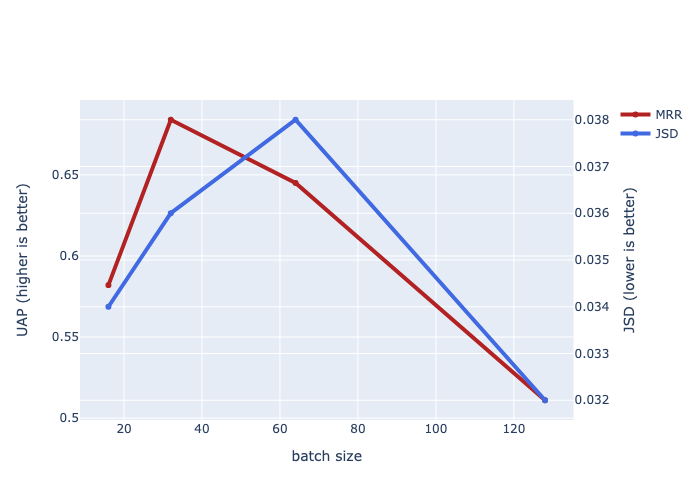}
    \caption{Effect of different batch sizes on M3L-Contrast trained on multilingual text data.} 
    \label{fig:text-only-batch-sizes}
\end{figure}

\begin{figure}[t!]
    \centering
    \includegraphics[width=0.55\textwidth]{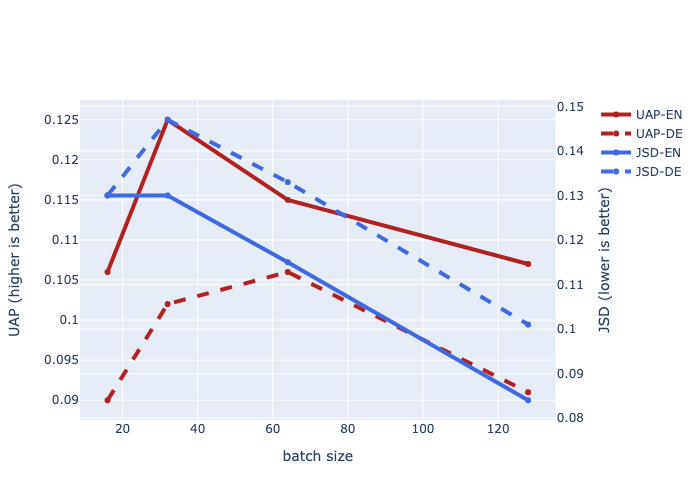}
    \caption{Effect of different batch sizes for M3L-Contrast trained on multilingual text \textit{and} images.} 
    \label{fig:text-images-batch-sizes}
\end{figure}

\begin{table*}[t!]
\centering
\begin{tabular}{l l | r r | r r | r r}
\hline
\multicolumn{2}{c}{\textbf{Encoders}} & \multicolumn{2}{c}{\textbf{EN-DE text}} &  \multicolumn{2}{c}{\textbf{EN-images}} & \multicolumn{2}{c}{\textbf{DE-images}}\\
\textbf{Text} & \textbf{Image}&  \textbf{MRR$\uparrow$} & \textbf{JSD$\downarrow$} & \textbf{UAP$\uparrow$} & \textbf{JSD$\downarrow$}  & \textbf{UAP$\uparrow$} & \textbf{JSD$\downarrow$} \\
\hline 
CLIP & CLIP & {0.613} & {0.035} & \textbf{0.125} & {0.130} & {0.102} & {0.147}\\
\hline
multilingual SBERT & CLIP & \textbf{0.716} & \textbf{0.029} & {0.119} & {0.137} & \textbf{0.114} & {0.147}\\
monolingual SBERTs & CLIP & {0.407} & {0.052} & {0.118} & \textbf{0.129} & {0.102} & \textbf{0.141}\\
multilingual SBERT & ResNet & {0.659} & {0.028} & {0.053} & {0.160} & {0.047} & {0.167}\\
monolingual SBERTs & ResNet & {0.347} & {0.050} & {0.053} & {0.145} & {0.052} & {0.157}\\
\hline
\end{tabular}
\caption{Effect of different encoder combinations for M3L-Contrast trained on \textbf{multilingual text and images} compared to CLIP with 100 topics.}
\label{tab:text-images-encoders}
\end{table*}

\subsection{Topic numbers}
In Figure~\ref{fig:text-only-topic-numbers} we show the models' performance on cross-lingual text retrieval (MRR) for $[25, 50, 100]$ topics. ZeroshotTM performs best for all topic numbers followed by M3L-Contrast. Figure~\ref{fig:text-images-topic-numbers} shows the results for text-image retrieval (UAP). M3L-Contrast performs best for all topic numbers while ZeroshotTM performs worst. In general, performance improves as the topic number increases.


\subsection{Batch sizes for M3L-Contrast}
We check batch sizes $[16, 32, 64, 128]$. Figure~\ref{fig:text-only-batch-sizes} shows the effect of increasing batch sizes for M3L-Contrast trained only on multilingual articles. We find that batch size 32 is the best for the multilingual setting. We also run similar experiments with multilingual text \textit{and} images (Figure~\ref{fig:text-images-batch-sizes}). In the multimodal setting, size 32 performs the best when English articles are matched to images while 64 is best for German. This is why we used batch size 32 for our experiments with M3L-Contrast. 


\subsection{Encoder combinations for M3L-Contrast}

To show how unaligned encoders perform with M3L-Contrast, we experiment with two encoder combinations: (1) a multilingual text encoder and an unaligned image encoder; and (2) unaligned monolingual text encoders and an unaligned image encoder. For this experiment, we train M3L-Contrast with multilingual articles \textit{and} images and evaluate the models on cross-lingual retrieval and text-image retrieval. Encoder details are in the Appendix. Results are shown in Table~\ref{tab:text-images-encoders}. 

For cross-lingual text retrieval, using a multilingual encoder performs best since this model is trained specifically on multilingual texts and has a larger embedding dimension than CLIP (768 and 512, respectively). For English text-image retrieval, it is expected that CLIP is the best since the text and image embeddings are already aligned (first row). CLIP image embeddings performed better than ResNet on all measures.

It is encouraging that M3L-Contrast with unaligned text and image embeddings still outperform ZeroshotTM and ZeroshotTM-KD (compare with Table~\ref{tab:results-text-images}, top part) even though those models use aligned embeddings. This shows that contrastive learning is effective in mapping unanligned embeddings into a shared topic space and that it is not necessary to use aligned embeddings in multimodal topic modelling.

\section{Conclusion}

We presented M3L-Contrast, a multimodal and multilingual neural topic model based on ZeroshotTM that uses pretrained document and image embeddings. M3L-Contrast is trained jointly on multilingual texts and images and does not require aligned embeddings. Since it is a \textit{multilingual} topic model it produces aligned language-specific topics. As a \textit{multimodal} topic model, it maps texts and images into a shared topic space and infers textual representations, through the topic words, of the semantic concepts present in the images. 



We show that in the multilingual setting, M3L-Contrast improves on PLTM, a classical multilingual topic model, and that it is competitive with ZeroshotTM in the alignment of topic distributions for comparable documents in different languages. In the multimodal setting, our model significantly improves on ZeroshotTM 
in aligning comparable texts and images in the topic space. Moreover, with unaligned text and image embeddings our model still performs better than ZeroshotTM that uses aligned embeddings.



Our proposed architecture can easily be extended to include other modalities beyond image and text. We also believe that M3L-Contrast will be useful in a low-resource setting, where aligned embeddings can be difficult to obtain.

\bibliographystyle{acl_natbib}
\bibliography{anthology,acl2021,custom}

\newpage
\section*{Appendix}

\subsection*{Data preprocessing}
We follow the training data preprocessing of \citet{bianchi-etal-2021-cross} for the BoW input: removing stopwords and retaining the 2000 most frequent words of each language as our vocabularies. We use the English and German stopword lists from NLTK\footnote{\url{https://www.nltk.org/}}.

\subsection*{Hyperparameters}
The neural topic models are trained on a single Nvidia V100 GPU (35 minutes) while PLTM is trained on a single Intel Xeon CPU (3 hours). During testing, we averaged the inferred topic distributions for each article/image from 20 samples. For all the neural models we used Adam optimizer with a learning rate of $2^{-3}$. We use a batch size of 64 except for M3L-Contrast. For M3L-Contrast, we set the temperature $\tau$ to 0.07 following \cite{guo2022multilingual}. We set the contrastive weight $s$ to 50 based on initial experiments. Tuning $\tau$ and $s$ are saved for future work.

\subsection*{Inference network}
We use the same inference network structure as ZeroshotTM~\cite{bianchi-etal-2021-cross}: one fully-connected hidden layer followed by softplus layer with 100 dimensions. We save the investigation of other inference network structures for future work.

\subsection*{Encoder Details}
We use SentenceBERT to encode all our data~\cite{reimers-gurevych-2020-making}~\footnote{\url{https://www.sbert.net/docs/pretrained_models.html}}. For a fairer comparison, we set the maximum sequence length of all text encoders to 128 tokens. The multilingual text encoder is \textit{paraphrase-multilingual-mpnet-base-v2}. For the monolingual encoders, the English encoder is \textit{all-mpnet-base-v2} and the German encoder is \textit{T-Systems-onsite/erman-roberta-sentence-transformer-v2}. ResNet embeddings are provided in this Kaggle challenge: \url{https://www.kaggle.com/competitions/wikipedia-image-caption}.

\subsection*{Potential impact and risks}
Our models are currently for research purposes only. We do not advise that it be used in production settings. Our models might associate images of people and objects with negative and insensitive stereotypes if the training data has these associations. Since we use CLIP to encode texts and images in our experiments, our models might also perpetuate the harmful stereotypes found in the CLIP training data discussed in~\cite{birhane2021multimodal}. The same issue applies to the other pretrained encoders we use in our experiments.





\end{document}